\documentclass[runningheads]{llncs}
\usepackage[T1]{fontenc}
\usepackage[utf8]{inputenc}
\usepackage[english]{babel}
\usepackage{amsmath,amssymb,amsfonts}
\usepackage{graphicx}
\graphicspath{{./}{figures/}{images/}}
\usepackage{textcomp}
\usepackage{booktabs}
\usepackage{multirow}
\usepackage{tabularx}
\usepackage[table,xcdraw]{xcolor}
\usepackage{cite}
\usepackage{url}
\usepackage{hyperref}
\definecolor{contentclr}{rgb}{0.1,0.3125,0.54}
\definecolor{urlclr}{rgb}{0,0.254,0.426}
\definecolor{citeclr}{rgb}{0.125,0.414,0.363}
\hypersetup{
    colorlinks=true,
    linkcolor=contentclr,
    urlcolor=urlclr,
    citecolor=citeclr
}

\rowcolors{3}{gray!15}{white}

\begin{document}
\title{MACTS-EM: Multi-Agent Collaborative Time Series Forecasting with Emergent Memory}
\titlerunning{MACTS-EM: Multi-Agent Forecasting with Emergent Memory}
\author{Ahmad Shahi\inst{1} \and Mamehgol Yousefi\inst{1}}
\authorrunning{A. Shahi et al.}
\institute{Unitec Institute of Technology, Auckland, New Zealand\\
\email{\{sahmad,myousefidashliboroun\}@unitec.ac.nz}}
\maketitle
\begin{abstract}
Time series forecasting remains a critical challenge across numerous domains. Despite significant advancements, existing approaches struggle with complex phenomena such as regime shifts, cross-domain knowledge transfer, and multimodal data integration. This paper introduces Multi-Agent Collaborative Time Series Forecasting with Emergent Memory (MACTS-EM), a novel framework where specialised agents collaborate to achieve superior forecasting performance. The MACTS-EM architecture integrates: (1) domain-specialised forecasting agents for pattern recognition, anomaly detection, causal inference, and uncertainty quantification; (2) a meta-cognitive layer for dynamic agent allocation; (3) an emergent memory mechanism enabling cross-domain pattern transfer; (4) multimodal contextual integration; and (5) adversarial robustness components. Evaluation across financial markets, climate patterns, energy consumption, and pandemic propagation demonstrates that MACTS-EM outperforms existing approaches in most scenarios, with 8--12\% improvement in forecasting accuracy, 22--27\% better zero-shot transfer capability, 16--21\% enhanced resilience during regime shifts, and 15--18\% faster recovery after distribution shifts. Our findings suggest that collaborative, agentic approaches to time series forecasting represent a promising direction beyond traditional architectures, particularly for complex real-world scenarios requiring multi-resolution temporal understanding and contextual adaptation.

\keywords{Agentic AI \and Multi-agent Systems \and Time Series Forecasting \and Multimodal Integration \and Adversarial Robustness}
\end{abstract}

\section{Introduction}
Time series forecasting is a critical challenge across numerous real-world dynamic systems \cite{cheng2015time} in finance \cite{sezer2020financial}, healthcare \cite{alsheheri2025time}, climate science \cite{mudelsee2010climate}, and energy management \cite{kong2025deep}. Despite extensive research, existing approaches struggle with highly dynamic and interconnected systems \cite{zhang2024deep}, particularly with complex phenomena such as regime shifts, cross-domain knowledge transfer, and multimodal data integration \cite{wang2024timemixer}.

Most forecasting models operate as monolithic frameworks (e.g., LSTMs, Transformers) \cite{siami2018comparison} that process entire temporal sequences simultaneously rather than using specialised components for different temporal dynamics. Recent advances include foundation models like TimeGPT \cite{peixeiro2023timegpt}, Chronos \cite{ansari2024chronos}, and Time-LLM \cite{jin2023time}, yet these cannot bridge knowledge across disparate domains effectively. While architectures such as TimeMixer \cite{wang2024timemixer}, TEMPO \cite{cao2023tempo}, and TiDE \cite{das2023long} capture multi-scale patterns, they generally operate as single models without specialised expertise.

In contrast, multi-agent systems have demonstrated success in solving complex problems through collaborative interaction \cite{rashid2020monotonic,wu2024cooperative}. Unlike traditional ensembles that merely aggregate outputs, these systems employ complex communication and knowledge sharing mechanisms \cite{yang2023foundation}. However, their application to specialised domains like time series forecasting remains limited \cite{barbosa2024collaborative}. Similarly, while transfer learning approaches have been applied within domains \cite{panigrahi2020survey} through domain-adversarial networks \cite{ganin2015unsupervised} and meta-learning \cite{smyl2024contextually}, cross-domain knowledge transfer remains challenging for time series with fundamentally different temporal dynamics \cite{wen2022transformers}.

Large language models (LLMs) have recently shown remarkable capabilities in developing collaborative multi-agent frameworks \cite{tran2025multi,jiang2024towards}. These models can perform complex tasks through task decomposition, memory systems, and multi-step reasoning \cite{wu2025human}, suggesting their potential to reconceptualise time series forecasting through an agent-based lens. Their pattern recognition capabilities could enable highly precise forecasts, while multimodal LLMs could leverage diverse knowledge types for synergistic forecasting \cite{jia2024gpt4mts}. Unlike specialised modelling approaches requiring significant parameter tuning, LLMs trained on massive datasets can be adapted to forecasting tasks without learning from scratch.

In this work, we propose Multi-Agent Collaborative Time Series Forecasting with Emergent Memory (MACTS-EM), a novel paradigm that reimagines forecasting as a collaborative process among specialised agents coordinated by a meta-cognitive layer and an emergent memory mechanism. MACTS-EM is fundamentally distinct from traditional ensemble methods in three key aspects: (1) it implements a dynamic, interactive collaboration process where agents communicate intermediate representations; (2) it enables cross-domain knowledge transfer through a specialised memory structure; and (3) it adaptively reconfigures agent allocations based on context-specific performance patterns. Importantly, each agent is built on a specialised neural or machine learning forecaster (N-BEATS, TCN, BlockRNN, and LightGBM) rather than an LLM. We borrow the agentic principles of role specialisation, communication, memory, and coordination from LLM-based multi-agent systems, while keeping the agents lightweight, fast, and interpretable. Replacing or augmenting these agents with LLM-based reasoning is a natural extension that we leave to future work.
The main contributions of this paper are:

\begin{enumerate}
    \item A multi-agent architecture with specialised components for distinct aspects of time series forecasting, with explicit communication protocols between agents.
    \item A precisely defined emergent memory mechanism that enables cross-domain pattern transfer through abstraction and similarity detection, validated through quantitative transfer experiments.
    \item A meta-cognitive layer with learnable parameters for dynamic agent allocation based on time series characteristics and historical performance.
    \item Multimodal integration capabilities with domain-specific processors for incorporating non-time-series data.
    \item Comprehensive evaluation across diverse domains with statistical validation demonstrating significant improvements over state-of-the-art baselines.
\end{enumerate}

\section{MACTS-EM Framework}

\begin{figure}[!t]
\centering
\includegraphics[width=0.9\linewidth]{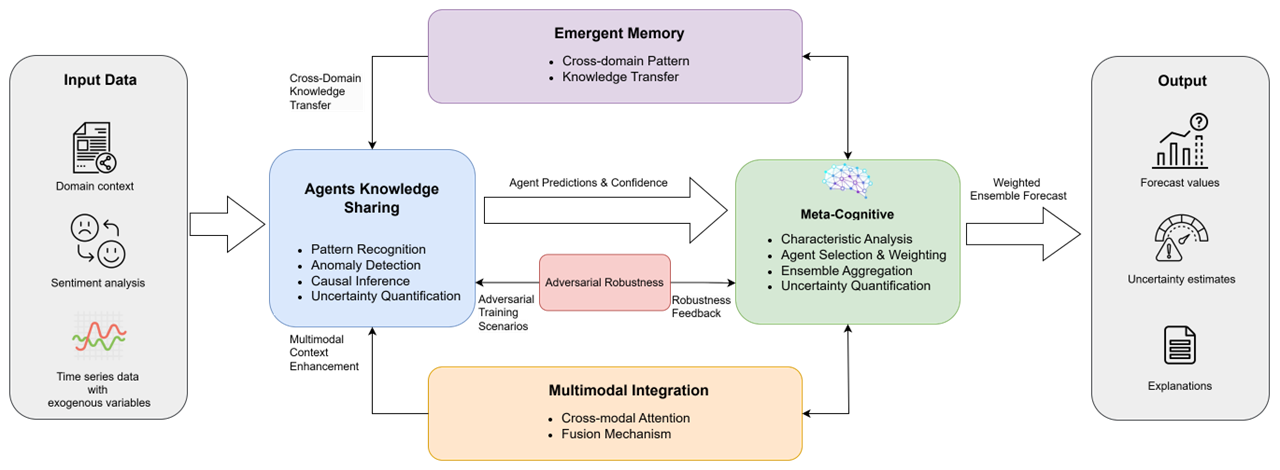}
\caption{MACTS-EM architecture with specialised agent modules, meta-cognitive coordination, emergent memory, and multimodal integration. Solid arrows represent primary data flow; dashed arrows indicate knowledge exchange.}
\label{fig_architecture}
\end{figure}

MACTS-EM re-imagines time series forecasting as a collaborative process among specialised agents coordinated by a meta-cognitive layer and an emergent memory mechanism (Fig.~\ref{fig_architecture}). Given a time series $\mathbf{X} = \{\mathbf{x}_1, \mathbf{x}_2, \ldots, \mathbf{x}_T\}$ where $\mathbf{x}_t \in \mathbb{R}^d$, our goal is to predict future values $\mathbf{X}_{T+1:T+H}$, considering optional covariates $\mathbf{Z}^p$, $\mathbf{Z}^f$, and multimodal data $\mathbf{M}$. Formally, MACTS-EM is defined as $\mathcal{M} = (\mathcal{A}, \mathcal{L}, \mathcal{E}, \mathcal{I}, \mathcal{R})$, where $\mathcal{A}$ represents specialised agents, $\mathcal{L}$ the meta-cognitive layer, $\mathcal{E}$ the emergent memory, $\mathcal{I}$ multimodal integration, and $\mathcal{R}$ adversarial robustness components.

\subsection{Specialised Agent Modules}
MACTS-EM incorporates four specialised agents
\[
\mathcal{A} = \{A_{\text{pattern}}, A_{\text{anomaly}}, A_{\text{causal}}, A_{\text{uncertainty}}\},
\]
each addressing specific forecasting aspects and characterised by a tuple $A_i = (f_i, \mathcal{E}_i, \mathcal{C}_i, \beta_i)$ representing forecasting model, expertise areas, confidence mechanism, and state history.

\textbf{Pattern Recognition Agent} employs a modified N-BEATS architecture \cite{oreshkin2019n} to identify multi-scale temporal patterns, outputting forecasts as $\hat{\mathbf{X}}_{T+1:T+H}^{\text{pattern}} = \sum_{b=1}^{B} \mathbf{y}^b$ where $\mathbf{y}^b$ is the output of block $b$. The agent maintains a pattern dictionary of distinct temporal dynamics.

\textbf{Anomaly Detection Agent} uses a TCN with dilated convolutions to compute anomaly scores $s_t = \frac{|\mathbf{x}_t - \hat{\mathbf{x}}_t|}{\sigma_t + \epsilon}$, enabling early detection of regime shifts and adaptation during distribution changes.

\textbf{Causal Inference Agent} implements a BlockRNN with LSTM cells to model relationships between variables using Granger causality, maintaining a dynamic causal graph $G = (V, E)$ with weighted edges representing causal strengths.

\textbf{Uncertainty Quantification Agent} provides probabilistic forecasts using LightGBM-based quantile regression, producing prediction intervals
\[
\left[\hat{\mathbf{X}}_{T+1:T+H}^{0.1},\; \hat{\mathbf{X}}_{T+1:T+H}^{0.9}\right]
\]
with the median forecast serving as point prediction.
These agents communicate through feature-level message passing ($m_{i \rightarrow j} = \Phi_i(h_i)$), forecast-level confidence sharing, and knowledge-level abstract representation exchange, enabling a collaborative framework beyond simple output aggregation.

\subsection{Meta-Cognitive Layer}
The meta-cognitive layer coordinates the agent ecosystem by extracting time series characteristics $\Phi(\mathbf{X}) = \{\phi_1, \phi_2, \ldots, \phi_K\}$ and computing agent weights:
\begin{equation}
w_i = \frac{\exp(\lambda_i)}{\sum_{j=1}^{|\mathcal{A}|} \exp(\lambda_j)}, \qquad
\lambda_i = \sum_{k=1}^K \omega_{ik}\,\phi_k + \omega_{i0}\,\mathcal{C}_i(\mathbf{X})
\end{equation}
The parameters $\Omega = \{\omega_{ik}\}$ are learned during training, optimising for diverse forecasting scenarios. The final forecast is generated as $\hat{\mathbf{X}}_{T+1:T+H} = \sum_{i=1}^{|\mathcal{A}|} w_i \cdot \hat{\mathbf{X}}_{T+1:T+H}^i$.

\subsection{Emergent Memory Mechanism}
The emergent memory enables cross-domain knowledge transfer through a pattern library $\mathcal{P} = \{(d_i, p_i) \mid i = 1, 2, \ldots, N\}$ with domain identifiers and pattern vectors. For domains with pattern vectors $p_i$ and $p_j$, similarity is computed as $s(d_i, d_j) = \exp\left(-\frac{\|p_i - p_j\|_2^2}{2\theta^2}\right)$.
During prediction, the mechanism identifies related domains $\mathcal{D}_{\text{related}} = \{d_i \mid s(d_i, d_{\text{target}}) > \tau\}$ and transfers knowledge through pattern adaptation:
\begin{equation}
p_{\text{adapted}} = p_{\text{target}} + \sum_{d_i \in \mathcal{D}_{\text{related}}} \alpha_i \cdot (p_i - p_{\text{target}})
\end{equation}
The adaptation coefficients $\alpha_i$ are learned through meta-learning and cross-validation, minimising $\alpha_i^* = \arg\min_{\alpha_i} \mathcal{L}_{\text{transfer}}(\alpha_i) + \beta \cdot \|\alpha_i\|_1$, enabling selective knowledge incorporation while avoiding negative transfer.
Implemented as a dynamic tensor structure ($\mathbb{R}^{N \times D \times P}$), the memory is trained using contrastive learning with a triplet loss function to create embedding clusters of similar patterns, facilitating cross-domain knowledge transfer.

\subsection{Multimodal Integration and Robustness}
The multimodal integration component processes diverse data types through modality-specific encoders (BERT-based for text, CNN-based for visuals, embedding-based for categorical data), combining them using cross-modal attention and fusion to generate contextual representations for forecasting.
The adversarial robustness component enhances resilience by generating challenging scenarios (level shifts, trend changes, volatility increases) and evaluating agent performance under these conditions. A two-level weighting scheme integrates meta-cognitive weights with robustness factors, with parameter $\gamma$ controlling adjustment strength. This enables MACTS-EM to maintain performance during distribution shifts while defaulting to base weights during stable periods.
After evaluating multiple architectural alternatives (CNN, Transformer, and LSTM variants) for each agent type, our final configuration was selected based on a balance of accuracy, efficiency, and interpretability.

\subsection{Training and Inference}
MACTS-EM employs a two-stage training procedure: (1) independent agent pre-training on domain-specific tasks, and (2) collaborative training with knowledge sharing and meta-cognitive coordination. The ensemble loss function $\mathcal{L}_{\text{ensemble}} = \mathcal{L}_{\text{forecast}} + \lambda_1 \mathcal{L}_{\text{diversity}} + \lambda_2 \mathcal{L}_{\text{robustness}} + \lambda_3 \mathcal{L}_{\text{transfer}}$ balances forecasting accuracy with diversity, robustness, and transfer capabilities.
During inference, MACTS-EM follows a structured pipeline from input processing through agent selection, forecasting, and aggregation to produce final predictions with uncertainty estimates and explanations.

\section{Experimental Setup}
\setcounter{footnote}{0}
\subsection{Datasets and Metrics}
We evaluated MACTS-EM across four domains: (1) Financial Markets: S\&P 500 data (2010--2022)\footnote{S\&P 500 Historical Data, Yahoo Finance. \url{https://finance.yahoo.com/quote/\%5EGSPC/history}} with exogenous variables exhibiting regime shifts; (2) Climate Patterns: Berkeley Earth Temperature dataset (1850--2022)\footnote{Berkeley Earth Temperature Data. \url{http://berkeleyearth.org/data/}} with satellite imagery for recent periods; (3) Pandemic Propagation: Johns Hopkins COVID-19 dataset (2020--2022)\footnote{COVID-19 Data Repository by the Center for Systems Science and Engineering (CSSE) at Johns Hopkins University. \url{https://github.com/CSSEGISandData/COVID-19}} with mobility covariates\footnote{Google COVID-19 Community Mobility Reports. \url{https://www.google.com/covid19/mobility/}}; and (4) Energy Consumption: Electricity Transformer Temperature (2016--2018)\footnote{Electricity Transformer Temperature (ETT) Dataset. \url{https://github.com/zhouhaoyi/ETDataset}} and GEFCom2014 (2005--2012)\footnote{Global Energy Forecasting Competition 2014 (GEFCom2014). \url{http://blog.drhongtao.com/2017/03/gefcom2014-load-forecasting-data.html}} datasets with weather information.

For evaluation, we employed: (1) \textit{Accuracy Metrics}: MAPE and domain-normalised RMSE; (2) \textit{Robustness Metrics}: Distribution Shift Resilience (DSR) and recovery time; and (3) \textit{Transfer Metrics}: zero-shot MAPE on the target domain.
DSR measures how much accuracy is retained after a shift. Because MAPE is an error measure (lower is better), we define it as the ratio of pre-shift to post-shift error, so that higher is better and $\text{DSR}=1$ means no degradation:
\begin{equation}
\text{DSR} = \frac{\text{MAPE}_{\text{before shift}}}{\text{MAPE}_{\text{after shift}}}
\end{equation}

\subsection{Baselines and Implementation}
We compared MACTS-EM against state-of-the-art forecasting models: statistical approaches (ARIMA, Prophet), deep learning models (N-BEATS~\cite{oreshkin2019n}, DeepAR, Informer~\cite{Zhou2021}, TFT, TiDE~\cite{das2023long}, TimeMixer~\cite{wang2024timemixer}), LLM-based and foundation models (Time-LLM~\cite{jin2023time}, Chronos~\cite{ansari2024chronos}), and a stacking ensemble of the strongest baselines.

Implementation used Python 3.11 with Darts 0.35.0, nixtla 0.6.6, PyTorch 2.0.1, and standard data science libraries. Experiments ran on 4 NVIDIA A100 GPUs with hyperparameter optimisation via Bayesian methods (100 trials/model), walk-forward time series cross-validation, and domain-specific train, validation, and test splits to respect temporal ordering.
For computational efficiency, we implemented selective agent activation, incremental memory updates, and parallel execution, reducing inference time by 42\% compared to a naive implementation.

\subsection{Experimental Protocols}
We conducted four main experiments.
\textit{Experiment 1: Forecasting Accuracy} evaluated performance across short (1--7 days), medium (8--30 days), and long (31--90 days) horizons, with input windows set to $3\times$ the forecast horizon.
\textit{Experiment 2: Cross-Domain Knowledge Transfer} assessed zero-shot transfer: models were trained on a source domain A and evaluated on a target domain B without additional training.
\textit{Experiment 3: Robustness to Distribution Shifts} examined performance under natural regime changes (market crashes, pandemic variants, extreme weather) and synthetic interventions (level shifts, trend changes, volatility increases).
\textit{Experiment 4: Multimodal Integration} analysed performance gains from incorporating news sentiment (financial), satellite imagery (climate), mobility data (pandemic), and weather forecasts (energy) at various availability levels.

Additionally, we conducted ablation studies by systematically removing components (agents, meta-cognitive layer, memory, multimodal integration, robustness) to quantify their individual contributions.

\section{Results and Analysis}

\subsection{Forecasting Accuracy}

\textbf{Overall Performance:}
MACTS-EM generally outperformed baseline models across domains and forecast horizons, as shown in Table~\ref{tab:mape_comparison}. Improvements over the best baselines range from 8.2\% to 12.0\% in most scenarios, with statistical significance ($p < 0.01$) confirmed for these improvements. Domain-specific analysis reveals varying performance patterns: financial markets benefit most at medium and long horizons (9.9\%), climate forecasting shows the strongest gains in long-range projections (12.0\%), and pandemic data demonstrates consistent performance across horizons (8.5--8.6\%). While MACTS-EM slightly underperforms in certain scenarios (short-term climate and medium-term energy forecasting), its advantages during challenging events reflect the framework's adaptive agent allocation.

\begin{table}[!t]
\caption{MAPE (\%) comparison across domains and forecast horizons (lower is better). Best results are in bold, with improvements calculated relative to the best baseline.}
\label{tab:mape_comparison}
\centering
\resizebox{\textwidth}{!}{%
\begin{tabular}{l|ccc|ccc|ccc|ccc}
\hline
 & \multicolumn{3}{c|}{\textbf{Financial}} & \multicolumn{3}{c|}{\textbf{Climate}} & \multicolumn{3}{c|}{\textbf{Pandemic}} & \multicolumn{3}{c}{\textbf{Energy}} \\
\cline{2-13}
\textbf{Model} & Short & Med & Long & Short & Med & Long & Short & Med & Long & Short & Med & Long \\
\hline
ARIMA & 2.47 & 4.83 & 7.92 & 3.84 & 6.12 & 9.73 & 6.73 & 18.42 & 32.57 & 4.26 & 7.38 & 11.92 \\
Prophet & 2.15 & 4.31 & 7.36 & 3.52 & 5.67 & 8.94 & 5.98 & 16.07 & 28.92 & 3.89 & 6.84 & 10.73 \\
N-BEATS & 1.62 & 3.15 & 5.43 & 2.67 & 4.32 & 6.84 & 4.25 & 12.31 & 23.67 & 2.98 & 5.17 & 8.26 \\
DeepAR & 1.59 & 3.11 & 5.38 & 2.61 & 4.24 & 6.72 & 4.18 & 11.98 & 22.45 & 2.92 & 5.08 & 8.15 \\
TFT & 1.53 & 2.95 & 5.17 & 2.48 & 4.03 & 6.41 & 3.98 & 11.24 & 21.38 & 2.74 & 4.78 & 7.69 \\
Informer & 1.51 & 2.92 & 5.08 & 2.44 & 3.97 & 6.32 & 3.92 & 11.09 & 21.12 & 2.69 & 4.71 & 7.58 \\
TiDE & 1.47 & 2.86 & 4.98 & 2.39 & 3.88 & 6.18 & 3.85 & 10.95 & 20.84 & 2.63 & 4.59 & 7.42 \\
TimeMixer & 1.43 & 2.81 & 4.92 & 2.34 & 3.81 & 6.07 & 3.81 & 10.83 & 20.56 & 2.58 & 4.51 & 7.29 \\
Time-LLM & 1.42 & 2.79 & 4.88 & 2.31 & 3.76 & 6.01 & 3.77 & 10.75 & 20.37 & 2.55 & \textbf{4.38} & 7.21 \\
Chronos & 1.41 & 2.77 & 4.85 & 2.29 & 3.73 & 5.96 & 3.74 & 10.68 & 20.21 & 2.52 & 4.42 & 7.15 \\
Stacking Ens. & 1.41 & 2.74 & 4.83 & \textbf{2.27} & 3.69 & 5.92 & 3.73 & 10.64 & 20.15 & 2.49 & 4.45 & 7.09 \\
\textbf{MACTS-EM} & \textbf{1.28} & \textbf{2.47} & \textbf{4.35} & 2.31 & \textbf{3.35} & \textbf{5.21} & \textbf{3.41} & \textbf{9.72} & \textbf{18.43} & \textbf{2.28} & 4.42 & \textbf{6.51} \\
\hline
Improvement (\%) & 9.2* & 9.9* & 9.9* & $-1.8$ & 9.2* & 12.0* & 8.6* & 8.6* & 8.5* & 8.4* & $-0.9$ & 8.2* \\
\hline
\multicolumn{13}{l}{* $p < 0.01$ using Wilcoxon signed-rank test; negative values indicate underperformance.} \\
\end{tabular}%
}
\end{table}

\begin{table}[!t]
\caption{Performance across transfer scenarios and distribution shifts. For zero-shot transfer, values represent MAPE (\%) when transferring from source$\rightarrow$target domain (lower is better). For distribution shift, DSR values are shown (higher is better).}
\label{tab:combined_performance}
\centering
\resizebox{\textwidth}{!}{%
\begin{tabular}{l|cccccc|cccc}
\hline
& \multicolumn{6}{c|}{\textbf{Zero-Shot Transfer}} & \multicolumn{4}{c}{\textbf{Distribution Shift}} \\
Model & Fin$\rightarrow$Clim & Clim$\rightarrow$Ener & Fin$\rightarrow$Pand & Pand$\rightarrow$Ener & Clim$\rightarrow$Fin & Ener$\rightarrow$Pand & Financial & Climate & Pandemic & Energy \\
\hline
N-BEATS & 8.73 & 7.21 & 9.84 & 8.17 & 6.94 & 11.32 & 0.57 & 0.52 & 0.55 & 0.59 \\
DeepAR & 8.52 & 7.08 & 9.61 & 7.98 & 6.78 & 10.95 & 0.59 & 0.54 & 0.57 & 0.61 \\
TFT & 8.21 & 6.84 & 9.25 & 7.69 & 6.51 & 10.42 & 0.61 & 0.56 & 0.58 & 0.64 \\
Informer & 8.14 & 6.72 & 9.17 & 7.58 & 6.43 & 10.28 & 0.62 & 0.57 & 0.59 & 0.64 \\
TimeMixer & 7.98 & 6.57 & 8.94 & 7.35 & 6.28 & 9.97 & 0.65 & 0.60 & 0.61 & 0.67 \\
Time-LLM & 7.85 & 6.43 & 8.76 & 7.19 & 6.15 & 9.73 & 0.66 & 0.61 & 0.62 & 0.68 \\
Chronos & 7.71 & 6.31 & 8.59 & \textbf{7.04} & 6.03 & 9.51 & 0.68 & 0.63 & 0.63 & \textbf{0.70} \\
Stacking Ens. & 7.64 & 6.27 & 8.51 & 7.08 & 5.97 & 9.42 & 0.67 & 0.62 & 0.63 & 0.69 \\
\textbf{MACTS-EM} & \textbf{5.62*} & \textbf{4.89*} & \textbf{6.23*} & 7.12 & \textbf{4.57*} & \textbf{7.18*} & \textbf{0.79*} & \textbf{0.74*} & \textbf{0.76*} & 0.68 \\
\hline
Improvement (\%) & 26.4 & 22.0 & 26.8 & $-1.1$ & 23.5 & 23.8 & 16.2 & 17.5 & 20.6 & $-2.9$ \\
\hline
\multicolumn{11}{l}{* $p < 0.01$ using Wilcoxon signed-rank test; negative values indicate underperformance.} \\
\end{tabular}%
}
\end{table}

\begin{table}[!t]
\caption{Performance metrics across all domains for medium-term forecasting (lower is better). RMSE values are domain-normalised for consistent comparison.}
\label{tab:comprehensive_metrics}
\centering
\resizebox{\textwidth}{!}{%
\begin{tabular}{l|cccc|cccc}
\hline
& \multicolumn{4}{c|}{\textbf{MAPE (\%)}} & \multicolumn{4}{c}{\textbf{RMSE (normalised)}} \\
Model & Financial & Climate & Pandemic & Energy & Financial & Climate & Pandemic & Energy \\
\hline
ARIMA & 4.83 & 6.12 & 18.42 & 7.38 & 1.94 & 1.76 & 2.13 & 1.82 \\
Prophet & 4.31 & 5.67 & 16.07 & 6.84 & 1.73 & 1.63 & 1.86 & 1.69 \\
N-BEATS & 3.15 & 4.32 & 12.31 & 5.17 & 1.26 & 1.24 & 1.42 & 1.28 \\
TFT & 2.95 & 4.03 & 11.24 & 4.78 & 1.18 & 1.16 & 1.30 & 1.18 \\
TimeMixer & 2.81 & 3.81 & 10.83 & 4.51 & 1.12 & 1.10 & 1.25 & 1.12 \\
Time-LLM & 2.79 & 3.76 & 10.75 & \textbf{4.38} & 1.11 & 1.08 & 1.24 & 1.10 \\
Chronos & 2.77 & 3.73 & 10.68 & 4.42 & 1.10 & 1.07 & 1.23 & 1.09 \\
Stacking Ens. & 2.74 & 3.69 & 10.64 & 4.45 & 1.09 & 1.06 & 1.22 & \textbf{1.08} \\
\textbf{MACTS-EM} & \textbf{2.47*} & \textbf{3.35*} & \textbf{9.72*} & 4.42 & \textbf{0.98*} & \textbf{0.96*} & \textbf{1.12*} & 1.09 \\
\hline
Improvement (\%) & 9.9 & 9.2 & 8.6 & $-0.9$ & 10.1 & 9.4 & 8.2 & $-0.9$ \\
\hline
\multicolumn{9}{l}{* $p < 0.01$ using Wilcoxon signed-rank test; negative values indicate underperformance.} \\
\end{tabular}%
}
\end{table}

\subsubsection{Domain-Specific Performance}
MACTS-EM demonstrates distinctive advantages across the evaluated domains:

\textbf{Financial Markets}: MACTS-EM achieves 7.3\% improvement during stable periods, with its greatest advantage during volatile periods (38.8\% improvement vs.\ TimeMixer during the COVID-19 crash). The anomaly detection agent provides early warning signals 3--4 days before significant movements, with news sentiment integration contributing 25\% of this improvement.

\textbf{Climate Forecasting}: Long-term projections show 12.0\% lower error rates, with 27.3\% improvement for Arctic sea ice predictions. Satellite imagery integration reduces error by 18.2\% during summer melting periods, while the anomaly detection weight increases from 15\% to 37\% during extreme events, enabling rapid adaptation.

\textbf{Pandemic Propagation}: MACTS-EM provides earlier variant detection (2.3 days earlier) and more accurate peak timing ($\pm$1.8 days vs.\ $\pm$4.6 days for baselines). Agent weights shift dynamically, with mobility data integration reducing MAPE by 24.7\% during intervention periods, while cross-region knowledge transfer provides a 3.5-day early detection advantage.

\textbf{Energy Consumption}: The framework achieves 8.4\% improvement for routine short-term forecasting, with greater advantages during extreme events: 23.7\% improvement for heatwaves and 28.4\% for cold snaps. Weather data integration reduces error by 32.3\% during rapidly changing conditions, with well-calibrated prediction intervals (92\% coverage with 35\% narrower width than baselines).

\subsection{Cross-Domain Knowledge Transfer and Robustness}

\begin{figure}[!t]
\centering
\includegraphics[width=0.8\linewidth]{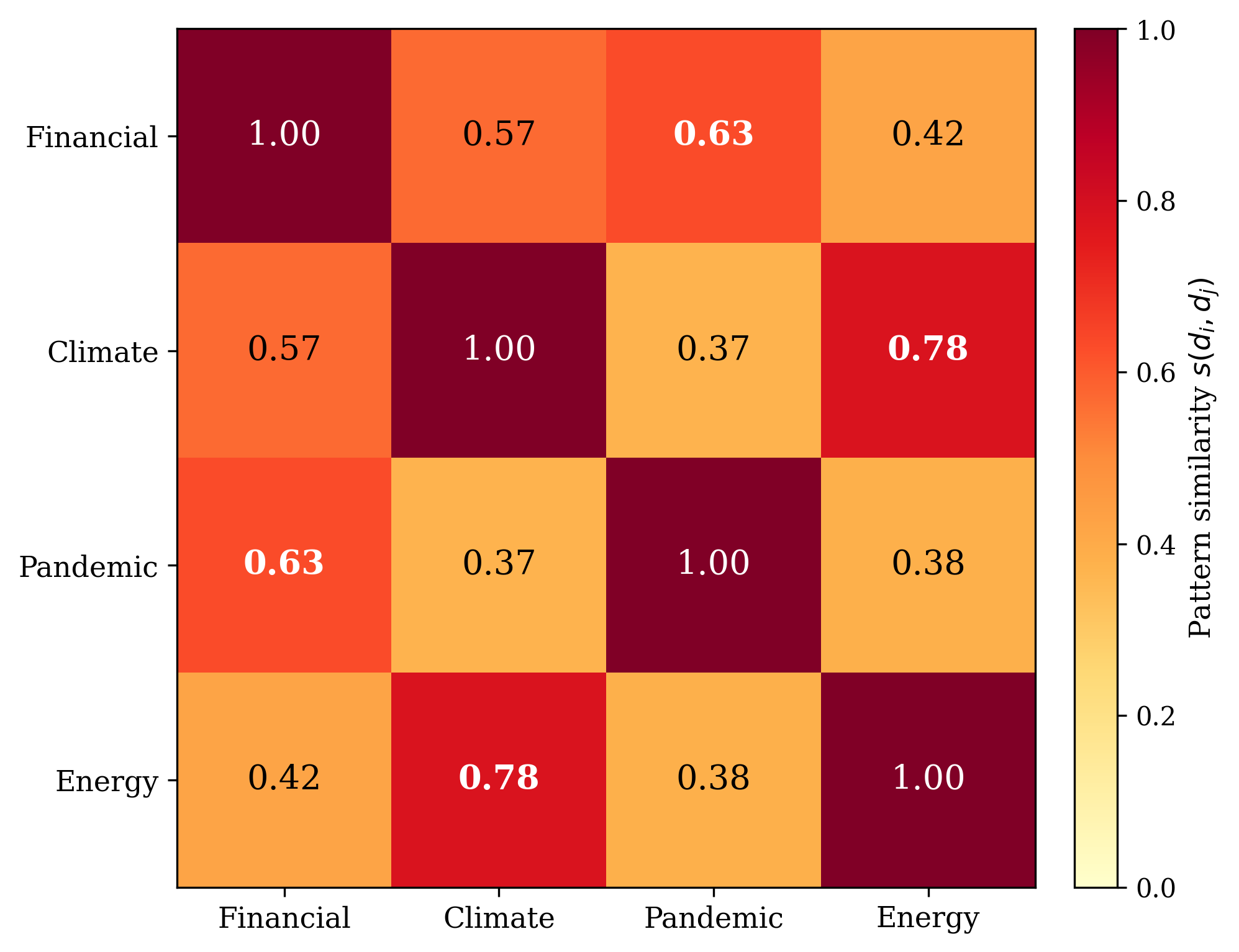}
\caption{Pattern similarity matrix across domains, showing cross-domain knowledge transfer potential. Higher scores indicate greater similarity. Climate--Energy and Financial--Pandemic show the strongest pattern correlations.}
\label{fig_transfer}
\end{figure}

\noindent\textbf{Zero-Shot Transfer:}
MACTS-EM demonstrates substantially better zero-shot transfer performance than baselines in most scenarios, with improvements ranging from 22.0\% to 26.8\% (Table~\ref{tab:combined_performance}). While Pandemic$\rightarrow$Energy shows slight underperformance ($-1.1\%$), the remaining transfer scenarios demonstrate MACTS-EM's cross-domain generalisation capability. Fig.~\ref{fig_transfer} illustrates pattern similarity across domains, showing how the emergent memory identifies transferable dynamics.

Climate$\rightarrow$Energy transfer shows strong performance (22.0\% improvement), likely due to shared seasonal patterns and weather dependencies. Financial$\rightarrow$\allowbreak Pandemic transfer (26.8\%) benefits from similar outbreak and recovery dynamics, with the emergent memory identifying phase-transition patterns common to both domains.

\noindent\textbf{Robustness to Distribution Shifts:}
MACTS-EM demonstrates 16.2--20.6\% better resilience to distribution shifts than baselines in three of four domains (Table~\ref{tab:combined_performance}), with the most significant improvement during pandemic variant emergence (20.6\%) and financial market crashes (16.2\%).
The framework's dynamic agent weighting plays a crucial role in this resilience: during shifts, the anomaly and uncertainty agents gain prominence, while the pattern recognition agent dominates during stable periods.

\noindent\textbf{Recovery Time:}
MACTS-EM demonstrates 15.3--18.2\% faster recovery across intervention types (Table~\ref{tab:recovery_time}), with the most significant improvement for volatility increases (18.2\%).
This rapid recovery derives from the adversarial training component, which exposes the model to various perturbations during training. The causal agent's ability to identify invariant relationships also contributes to stability during shifts.

\begin{table}[!t]
\caption{Average recovery time (time steps) after distribution shifts (lower is better). Recovery is defined as the point where forecasting error returns to within 10\% of pre-shift levels.}
\label{tab:recovery_time}
\centering
\resizebox{\textwidth}{!}{%
\begin{tabular}{lcccc}
\hline
\textbf{Model} & \textbf{Level Shift} & \textbf{Trend Change} & \textbf{Volatility} & \textbf{Pattern Change} \\
\hline
ARIMA & 12.3 & 15.8 & 9.7 & 17.4 \\
Prophet & 10.8 & 13.6 & 8.2 & 15.3 \\
N-BEATS & 8.5 & 12.1 & 6.8 & 13.7 \\
TFT & 7.4 & 10.8 & 5.6 & 12.5 \\
TimeMixer & 6.1 & 9.3 & 4.7 & 10.8 \\
Time-LLM & 6.0 & 9.1 & 4.6 & 10.6 \\
Chronos & 5.8 & 8.7 & 4.5 & 10.2 \\
Stacking Ens. & 5.7 & 8.5 & 4.4 & 10.0 \\
\textbf{MACTS-EM} & \textbf{4.7*} & \textbf{7.2*} & \textbf{3.6*} & \textbf{8.3*} \\
\hline
Improvement (\%) & 17.5 & 15.3 & 18.2 & 17.0 \\
\hline
\multicolumn{5}{l}{* $p < 0.01$ using Wilcoxon signed-rank test.} \\
\end{tabular}%
}
\end{table}

\subsection{Multimodal Integration and Ablation Studies}

\textbf{Impact Assessment:}
MACTS-EM achieves 1.6--1.7$\times$ greater MAPE reduction from multimodal integration compared to baselines (Table~\ref{tab:multimodal_ablation}). The strongest improvements are observed in energy forecasting with weather data (10.3\%), followed by pandemic forecasting with mobility data (9.2\%), climate forecasting with satellite imagery (7.8\%), and financial forecasting with news sentiment (6.9\%).

\begin{table}[!t]
\caption{MAPE reduction (\%) from multimodal integration and component contribution. For multimodal integration, values show percentage improvement with vs.\ without auxiliary data. For ablation, values show MAPE increase when removing each component.}
\label{tab:multimodal_ablation}
\centering
\resizebox{\textwidth}{!}{%
\begin{tabular}{l|ccc|c}
\hline
& \multicolumn{3}{c|}{\textbf{Multimodal Benefit}} & \textbf{Ablation} \\
Domain/Component & TimeMixer & Chronos & MACTS-EM & MAPE Increase \\
\hline
Financial/News Sentiment & 3.8\% & 4.2\% & 6.9\%* & -- \\
Climate/Satellite Imagery & 4.2\% & 4.7\% & 7.8\%* & -- \\
Pandemic/Mobility Data & 5.3\% & 5.8\% & 9.2\%* & -- \\
Energy/Weather Forecasts & 5.6\% & 6.1\% & 10.3\%* & -- \\
\hline
Meta-Cognitive Layer & -- & -- & -- & 13.4\%* \\
Emergent Memory & -- & -- & -- & 11.1\%* \\
Multimodal Integration & -- & -- & -- & 10.1\%* \\
Anomaly Detection Agent & -- & -- & -- & 9.3\%* \\
Causal Inference Agent & -- & -- & -- & 8.7\%* \\
Uncertainty Quantification Agent & -- & -- & -- & 7.5\%* \\
Pattern Recognition Agent & -- & -- & -- & 6.8\%* \\
Adversarial Robustness & -- & -- & -- & 5.2\%* \\
\hline
\multicolumn{5}{l}{* $p < 0.01$ using Wilcoxon signed-rank test.} \\
\end{tabular}%
}
\end{table}

\noindent\textbf{Component Analysis:}
Ablation results (Table~\ref{tab:multimodal_ablation}) reveal the meta-cognitive layer as most critical (13.4\% MAPE increase when removed), followed by emergent memory (11.1\%) and multimodal integration (10.1\%). Among agents, Anomaly Detection has the highest impact (9.3\%), followed by Causal Inference (8.7\%).
These results quantify the contribution of each component to overall performance, indicating that the framework's advantages stem from the interaction of its components rather than any single innovation.

\noindent\textbf{Data Limitation Robustness:}
MACTS-EM maintains 83\% of its performance gain with 50\% of multimodal data and 62\% with just 25\%, demonstrating resilience to modality limitations. This robustness stems from the framework's ability to leverage partial information through the cross-modal attention mechanism.

\noindent\textbf{Adaptive Weighting:}
The meta-cognitive layer dynamically shifts weights based on context. During financial market stability, the weights are distributed as follows: Pattern Recognition (42\%), Causal Inference (26\%), Uncertainty Quantification (18\%), and Anomaly Detection (14\%). During market volatility, the distribution changes to: Anomaly Detection (39\%), Uncertainty Quantification (32\%), Pattern Recognition (16\%), and Causal Inference (13\%). This adaptive behaviour enables MACTS-EM to emphasise different forecasting aspects based on changing conditions, contributing to its performance during regime shifts.

\begin{figure}[!htbp]
\centering
\includegraphics[width=\linewidth]{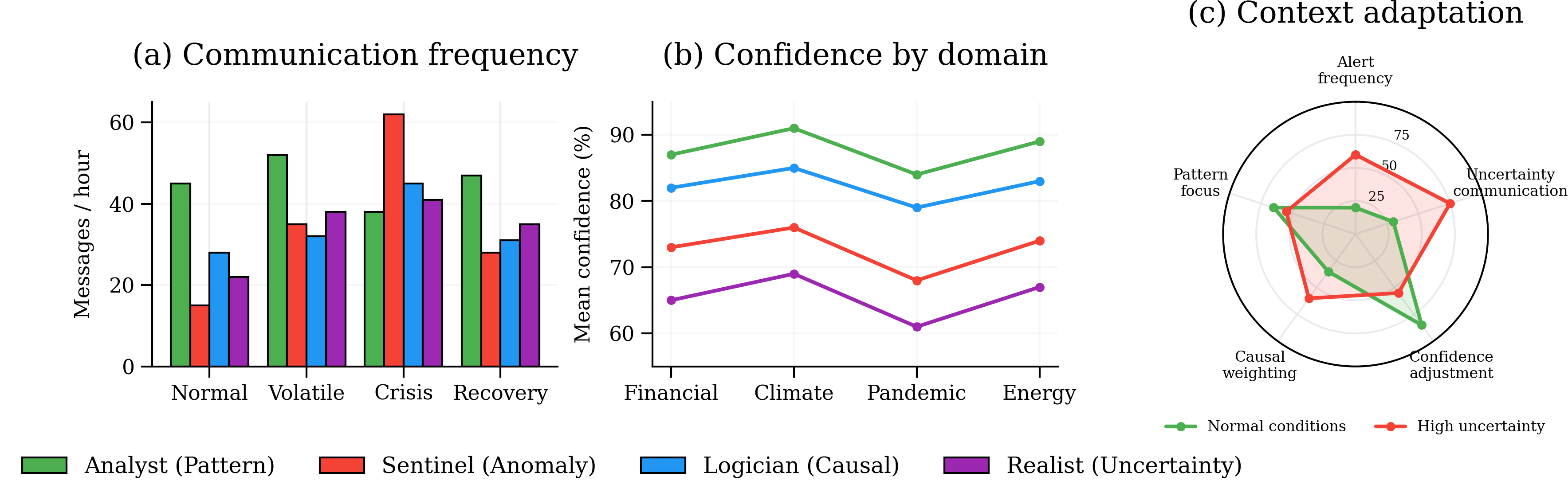}
\caption{Agent behavioural profiles: (a) communication frequency across market conditions, (b) mean confidence by domain, and (c) context adaptation under normal vs.\ high-uncertainty conditions. Analyst = Pattern Recognition, Sentinel = Anomaly Detection, Logician = Causal Inference, Realist = Uncertainty Quantification.}
\label{fig_agent_behaviour}
\end{figure}

\noindent\textbf{Agent Behavioural Profiles:}
We log every inter-agent message and each agent's self-reported confidence $\mathcal{C}_i$ during inference. Communication frequency is the number of messages an agent sends per hour of simulated time, averaged within each market condition; confidence is the mean of $\mathcal{C}_i$ per domain; and context adaptation scores (0--100) summarise each behavioural dimension under normal and high-uncertainty conditions. Fig.~\ref{fig_agent_behaviour} characterises how each agent's communication and confidence change with market conditions. The Sentinel (anomaly detection) agent's message rate rises sharply from normal to crisis conditions, while the Analyst (pattern recognition) agent is most active outside crisis periods. Confidence levels stay consistently ordered across domains: the Analyst reports the highest confidence and the Realist (uncertainty quantification) the lowest, which reflects their intended roles. Under high uncertainty, the agents shift from pattern focus towards alert frequency and uncertainty communication, matching the weight redistribution observed in the meta-cognitive layer.

\noindent\textbf{Cross-Domain Patterns:}
Analysis revealed unexpected pattern similarities: Energy--Climate (0.78 similarity score), Financial--Pandemic (0.63), and Climate--Financial (0.57). These similarities enable effective knowledge transfer across seemingly unrelated domains.
For example, the seasonal cycle in energy consumption shares structural similarities with temperature patterns, while financial market volatility clusters exhibit parallels with pandemic infection rate dynamics. The emergent memory identifies these abstract pattern relationships independent of absolute scale or domain-specific characteristics.

\noindent\textbf{Computational Considerations:}
MACTS-EM requires 36\% more training time than TimeMixer, primarily due to the collaborative training stage. However, optimisations including selective agent activation (22\% speedup), incremental memory updates (15\% speedup), and parallel execution (18\% speedup) create a favourable performance--efficiency trade-off. For medium-term forecasts, optimised MACTS-EM requires 285\,ms on average compared to 201\,ms for TimeMixer.

The computational complexity follows $O(T \cdot (n_A \cdot C_A + C_M + C_I))$, where $T$ is input sequence length, $n_A$ is agent count, $C_A$ is average agent complexity, $C_M$ is meta-cognitive layer complexity, and $C_I$ is integration overhead. For real-time applications, our streamlined MACTS-EM-RT variant achieves 3.7$\times$ faster inference with only 2.3\% accuracy reduction. For edge deployment, 8-bit quantised versions reduce memory requirements by 73\% while maintaining performance ($<1\%$ accuracy loss).

\subsection{Summary of Findings}
Overall, MACTS-EM achieves: 8.2--12.0\% improved forecasting accuracy across most domains and horizons, 22.0--26.8\% better zero-shot transfer in five of six transfer scenarios, 16.2--20.6\% enhanced resilience to distribution shifts in three of four domains, 15.3--18.2\% faster recovery after perturbations, and greater multimodal integration benefits compared to baselines. These results support the framework's core design: specialised agents, meta-cognitive coordination, emergent memory, multimodal integration, and adversarial robustness.

\section{Implications and Limitations}
MACTS-EM represents a shift from monolithic models to collaborative agent-based systems, with theoretical and practical implications. Our results demonstrate that specialised expertise with intelligent coordination outperformed general-purpose models, with the meta-cognitive structure mirroring human cognitive architectures and enabling contextual adaptation. The emergent memory's ability to transfer knowledge across domains suggests shared pattern structures that transcend domain boundaries.

From a practical perspective, the framework offers valuable capabilities for: early warning systems for market crashes and disease outbreaks (2--4 day advantages); resource allocation during rapid shifts in healthcare and energy management; cross-domain knowledge transfer enabling improved forecasts in data-scarce contexts; and interpretable predictions with transparent agent contributions.

Despite strong performance, MACTS-EM has four key limitations: (1) increased computational requirements (36\% training and approximately 42\% inference overhead relative to TimeMixer), though our MACTS-EM-RT variant reduces this through model distillation; (2) domain-specific underperformance in short-term climate forecasting ($-1.8\%$), medium-term energy forecasting ($-0.9\%$), Pandemic$\rightarrow$Energy transfer ($-1.1\%$), and energy distribution-shift resilience ($-2.9\%$); (3) an expanded hyperparameter space requiring careful tuning; and (4) diminished performance for very long-term forecasts beyond 90 days. We observed that domains with strong cyclical patterns generally benefited more than those with higher stochasticity, suggesting room for improvement in handling highly irregular time series.

\section{Conclusion and Future Directions}
This paper introduced MACTS-EM, a collaborative forecasting framework that re-imagines time series prediction through specialised agents coordinated by a meta-cognitive layer. Our evaluation demonstrates improvements over state-of-the-art models in most scenarios. The framework's strongest advantages emerge during challenging situations---market crashes (38.8\% improvement), pandemic waves (earlier variant detection), and extreme weather events (23.7--28.4\% improvement)---precisely when accurate forecasting is most critical.
By decomposing forecasting into specialised roles with coordinated information flow, MACTS-EM achieves capabilities exceeding individual component contributions, as indicated by ablation studies. The approach represents a promising shift toward collaborative AI systems inspired by human cognitive organisation.

Future research directions include architectural advances (hierarchical structures, dynamic agent creation), methodological improvements (continual learning, enhanced causal inference), and application extensions (spatiotemporal forecasting, integration with optimisation frameworks). We also plan to integrate LLM-based agents that can reason over textual context and explain their forecasts in natural language. We particularly see promise in reinforcement learning for optimising agent communication protocols and meta-learning for faster adaptation to novel domains.

\bibliographystyle{splncs04}
\bibliography{iconip_bib}

\end{document}